\documentclass{article}
\usepackage{graphicx} 
\usepackage{comment}
\usepackage{amsmath}

\title{Emotion Detection through Body Gesture and Face}
\author{Haoyang Liu}

\begin{document}

\maketitle

\section{Abstract}
The project leverages advanced machine and deep learning techniques to address the challenge of emotion recognition by focusing on non-facial cues, specifically hands, body gestures, and gestures. Traditional emotion recognition systems mainly rely on facial expression analysis and often ignore the rich emotional information conveyed through body language. To bridge this gap, this method leverages the Aff-Wild2 and DFEW databases to train and evaluate a model capable of recognizing seven basic emotions (angry, disgust, fear, happiness, sadness, surprise, and neutral) and estimating valence and continuous scales wakeup descriptor.

Leverage OpenPose for pose estimation to extract detailed body posture and posture features from images and videos. These features serve as input to state-of-the-art neural network architectures, including ResNet, and ANN for emotion classification, and fully connected layers for valence arousal regression analysis. This bifurcation strategy can solve classification and regression problems in the field of emotion recognition.

The project aims to contribute to the field of affective computing by enhancing the ability of machines to interpret and respond to human emotions in a more comprehensive and nuanced way. By integrating multimodal data and cutting-edge computational models, I aspire to develop a system that not only enriches human-computer interaction but also has potential applications in areas as diverse as mental health support, educational technology, and autonomous vehicle systems. 

\textit{Keywords – Aff-Wild2, DFEW, ResNet, DenseNet, ANN, Valence and Arousal}


\section{Introduction}
\subsection{Background}
Human-computer interaction (HCI) has expanded dramatically with the emergence of machine and deep learning technologies, aiming to create more natural and intuitive ways for machines to understand and respond to human needs. A key aspect of this interaction is emotion recognition, which is the process of recognizing human emotions through various means, including facial expressions, speech, and, as the focus of this project, body gestures. Historically, facial expressions have been the primary conduit for emotion recognition systems, exploiting the nuances of the human face to map to universally recognized basic emotions: anger, disgust, fear, happiness, sadness, surprise, and neutral states. However, this approach ignores the rich nonverbal cues humans display through body language. 

Recent explorations in the field of affective computing have shown that body gestures can tell as much, if not more, about a person's emotional state as facial expressions alone. These non-facial cues can provide additional context and improve the accuracy of emotion recognition systems, especially when facial expressions are ambiguous or social conditions mask true feelings. Therefore, emotion recognition is crucial in the application areas such as marketing, human–robot interaction, healthcare, mental health monitoring, and security \cite{kamble2023comprehensive}. The study of emotions for healthcare includes vast neurological disorders like sleep disorders \cite{dahl2007sleep}, evaluation of sleep quality \cite{mauss2013poorer}. 

This project aims to harness the untapped potential of body language as a medium for emotion recognition. By leveraging advanced machine and deep learning techniques, such as OpenPose for pose estimation, and complex neural networks, such as ResNet, DenseNet, and Visual Transformers (ViT), the project aims to classify of seven basic emotions. And estimate the valence and arousal emotion descriptors are derived from human body gestures. 

The databases that will be used, Aff-Wild2 \cite{kollias2019deep,kollias2019expression,kollias2022abaw,kollias2023abaww,kollias2023btdnet,kollias2023facernet,kollias2023multi,kollias20246th,kollias2024distribution,kollias2024domain,kolliasijcv,zafeiriou2017aff,hu2024bridging,psaroudakis2022mixaugment,kollias2020analysing,kollias2021distribution,kollias2021affect,kollias2019face,kollias2021analysing,kollias2023ai,kollias2023deep2,arsenos2023data,gerogiannis2024covid,jiang2020dfew,salpea2022medical,arsenos2024uncertainty,karampinis2024ensuring,arsenos4674579nefeli,miah2024can,arsenos2024commonn} and DFEW, provide rich datasets on which models can be trained to account for continuous emotional states, ranging from highly positive to very negative (valence), and from very passive to highly active (arousal). Classification of basic emotions will serve as a discrete problem-solving task, while valence arousal estimation will solve a more complex regression problem. 

\subsection{Problem Statement}
In the rapidly developing field of human-computer interaction, the accurate recognition and interpretation of human emotions plays a crucial role. Traditional emotion recognition systems focus primarily on facial expressions, which, while effective, cannot provide a complete picture of the human emotion spectrum. This limitation becomes especially apparent in situations where facial cues are invisible or influenced by social and cultural norms (like wear mask). Therefore, there is a huge gap in exploiting the full potential of non-verbal cues, especially body gestures, which are often overlooked in emotion recognition systems.

Furthermore, most existing emotion recognition models are build around discrete emotional states, mainly the seven basic emotions. However, human emotions are more subtle and often exist on a spectrum, especially when considering the dimensions of valence and arousal. The lack of models that accurately explain and quantify these dimensions limits the applicability of emotion recognition to domains that require a more complex understanding of human emotions, such as mental health assessment and adaptive learning systems. 

The challenge, therefore, is to develop a comprehensive emotion recognition system that goes beyond only facial analysis to incorporate the subtleties of body gesture to provide a more complete and accurate understanding of human emotions. Such a system must not only classify basic emotional states but must also leverage advanced machine and deep learning techniques to estimate complex emotion descriptors of valence and arousal. The effective integration of these technologies has the potential to revolutionize the field of affective computing and expand its applications in different fields.

\subsection{Aim}
The purpose of this project is to develop an emotion recognition system that can identify human emotions with high accuracy by analysing body gestures and facial expression. The system utilizes machine and deep learning methods and is designed to surpass the capabilities of traditional facial recognition models. It will cover the seven basic emotional states, valence and arousal. With this innovation, we seek to capture the subtlety and complexity of human emotions and promote better understanding and interaction within the field of affective computing.

\subsection{Objectives}
\begin{itemize}
    \item Pose Estimation Integration: To integrate OpenPose to accurately capture body gestures from visual data. It capable of delineating individual body parts and joint movements, which are critical for interpreting physical expressions related to emotional states.
    \item Model Development: To develop and train deep learning models, such as artificial neural networks (ANNs), ResNet, and Visual Transformers, for the classification of the seven basic emotions.
\item Emotion classification: To construct a classify model capable of 7 basic emotion states. Inputs are from face expressions and body gestures, through the model to determine the specific emotional state among the seven basic emotions.
\item Valence-Arousal Analysis: To construct a regression model capable of estimating valence and arousal levels, providing a mapping of emotional states.
\item Performance Evaluation: To rigorously evaluate the performance of the models using metrics such as accuracy, precision, and recall.
\item System Optimization: To fine-tune the system for real-time processing and ensure it is robust against variations in environmental conditions.

\end{itemize}

\section{Literature Review}
Emotion recognition has become a popular research topic in the past decade. While works based on facial expressions or speech abound, identifying the effects of body posture remains a less explored topic \cite{noroozi2018survey}, This chapter provides a detailed analysis of the impact of associate technology and provides information on existing technologies in place in order to gather requirements for the proposed solution. Then what lessons can be learned from these approaches?
\subsection{Emotion Recognition Techniques}
There are several traditional approaches to process emotion recognition. Like from speech and facial expressions are the most employed mechanisms for emotion identification among physical signals \cite{shu2018review}. After that, Physiological Response Measurement is a good approach to recognize emotion. For example, electroencephalogram (EEG), electrocardiogram (ECG), electromyogram (EMG), galvanic skin response (GSR), respiration (RSP), skin temperature, photoplethysmography, and eye tracking (ET) are the most employed physiological signals for emotion recognition \cite{shu2018review}. Among physiological signals, the most often utilized modalities for detecting human emotions are EEG, GSR, ECG, and ET \cite{khare2023emotion}. Furthermore, body gesture. As a crucial component of body language, also a good approach to recognize emotion.
\subsection{Emotion Recognition from Body Gesture and Pose}
Psychological research further shows that body gestures conveys nonverbal emotional cues, include some part face and voice cannot express. However, recognizing emotion based on body gestures remains less explored.

Emotion recognition from body gesture is a field of study in artificial intelligence and human-computer interaction focusing on the ability of machines to identify and interpret human emotions based on physical expressions and movements. This technology uses sensors, computer vision, and machine learning algorithms to analyse the posture, movement, and gestures of a person.

The Multi-view emotional expressions dataset using 2D pose estimation as featured in Nature emphasizes the potency of body expressions in conveying emotional shifts, sometimes even surpassing facial or vocal expressions in efficacy \cite{zhang2023multi}.

The exploration of body gesture as a component of body language in emotion recognition is less explored compared to face expression-based and speech-based approaches. However, the works such as those presented in IEEE Xplore suggest that deep learning can successfully be employed to recognize emotions from body gestures, using large datasets of multi-view RGB videos for training \cite{shen2019emotion}.

This burgeoning interest in body gesture and pose for emotion recognition signifies a pivotal shift towards more comprehensive and possibly more naturalistic modes of human affect analysis. As such, it holds considerable promise for applications in areas ranging from interactive gaming to psychological analysis, and from assistive technologies to security systems.
\subsection{Machine and Deep Learning in Emotion Recognition}
Machine learning is like teaching a computer to make decisions or predictions based on past examples. You give the computer data, and it learns patterns from that data. While deep learning is a special way of doing machine learning that involves layers of processing to understand complex patterns. Think of it like stacking several filters to refine what you learn from the data, each layer focusing on learning more detail. 

Neural Networks are the foundation of deep learning. They are computing systems vaguely inspired by the biological neural networks that constitute animal brains. An artificial neural network consists of a collection of connected units called neurons, which process data in layers. The first layer takes in the input data, and each subsequent layer builds a more abstract representation of the data, with the final layer producing the output. 

In recent years, the rapid development of machine learning (ML) and deep learning (DL) and information fusion has made it possible to give machines/computers the ability to understand, identify and analyse emotions. Emotion recognition has attracted increasing interest from researchers in different fields. Human emotions can be recognized through facial expressions, speech, behaviour (gestures/postures), or physiological signals. 

The SVM-based ML decision-making has been proven the most effective and preferred emotion recognition model. The ability of DL models to automatically extract and classify deep features is gaining popularity and has been increasing in the usage of CNN models \cite{khare2023emotion}. 

\subsection{OpenPose and Its Contributions}
OpenPose has revolutionized the field of human-computer interaction by providing a tool for real-time multi-person 2D pose estimation. Developed at Carnegie Mellon University, it has been instrumental in advancing research in various domains, including emotion recognition. The original work introducing OpenPose offered a method uses a nonparametric representation, which referred as Part Affinity Fields (PAFs), to learn to associate body parts with individuals in the image, allowing for effective pose estimation in complex scenes with multiple people \cite{cao2017realtime}. 

Because OpenPose can identify subtle body gestures and poses that are indicative of emotional states, its application in emotion recognition is compelling. Its use extends beyond the capabilities of facial emotion recognition, particularly in scenarios where the face may not be visible or expressions are subtle and thus harder to detect. For instance, research has employed OpenPose to analyse body gestures in conjunction with deep learning models for emotion recognition, a novel approach is proposed to use deep neural network fuse skeleton and RGB features only using single-modality RGB video data. Experimental results show them approach achieves substantial improvements both in individual categories and overall and is provided with stronger generalization capability as well \cite{shen2019emotion}. 

Based on the capability of OpenPose, I can get the keypoints of a person in picture. OpenPose can process each frames in a video, therefore I can get images from each video’s frames. Then use OpenPose to process each images to get the keypoints of person in the images. The keypoints will shows the body poses with lines with different colours and also detect the outline of their face. Thus I can get the images that contains the body pose with the face outlines and disable the original image just keep the keypoints, which will become dataset for training and validating. Figure 1 shows a example of processing image via OpenPose. 
\begin{figure}
    \centering
    \includegraphics[width=.5\linewidth]{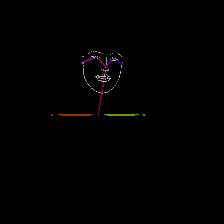}
    \caption{\textit{Figure 1 Processed Frame Image by OpenPose }}
\end{figure}

\subsection{Databases for Emotion Recognition}
The Aff-Wild2 database, an extension of the original Aff-Wild dataset, is the first to capture real-world affect in-the-wild. It offers a substantial volume of data, including videos download from Youtube with associated annotations for valence and arousal. The database's annotations of the seven basic emotions and continuous valence- arousal labels allow for both classification and regression problem formulations in emotion recognition tasks \cite{kollias2019deep,kollias2019expression,kollias2022abaw,kollias2023abaww,kollias2023btdnet,kollias2023facernet,kollias2023multi,kollias20246th,kollias2024distribution,kollias2024domain,kolliasijcv,zafeiriou2017aff,hu2024bridging,psaroudakis2022mixaugment,kollias2020analysing,kollias2021distribution,kollias2021affect,kollias2019face,kollias2021analysing,kollias2023ai,kollias2023deep2,arsenos2023data,gerogiannis2024covid,jiang2020dfew,salpea2022medical,arsenos2024uncertainty,karampinis2024ensuring,arsenos4674579nefeli,miah2024can,arsenos2024commonn,kollias20247th}. 

This dataset consisting of more than 500 videos, having 2,000,000 more frames, all of person in video which have been annotated. The videos in the dataset are captured in unconstrained settings, with varying illumination conditions, head poses, occlusions, and backgrounds, making it more challenging and realistic than controlled laboratory settings. 

DFEW (Dynamic Facial Expression in the Wild) is another prominent dataset that provides a collection of facial expressions captured in natural settings. With its large- scale, DFEW can be served as a benchmark for researchers to develop and evaluate their methods for dealing with dynamic facial expression recognition (FER) in the wild \cite{jiang2020dfew}. 

This dataset contains over 16000 videos from thousands of movies, these video clips contain various challenging interference in practical scenarios such as extreme illumination, occlusions, and capricious pose changes \cite{jiang2020dfew}. These videos will provide lots of images with different emotions. The videos in the DFEW dataset are collected from popular movies, which means that the facial expressions are captured in various unconstrained settings, with different illumination conditions, occlusions, and head poses. 

\subsection{Emotional States}
The classification of emotions into discrete categories, particularly the basic emotions of anger, disgust, fear, happiness, sadness, surprise, and neutral, has served as a foundational element in the field of affective computing. This traditional approach makes use of the classification of emotions into these seven discrete states in order to expedite the creation of computational models that are capable of identifying and interpreting human emotions. Supervised learning approaches, a subset of machine learning techniques, are ideally suited to this classification job. Supervised learning algorithms are best suited in scenarios where labelled datasets are abundant since they require a dataset containing labelled examples from which to learn. 

In the context of emotion recognition, datasets often consist of videos or images of individuals expressing various emotions. These expressions are annotated with labels that correspond to the basic emotions. For example, annotations are assigned numerical values from 1 to 7, each number representing a different basic emotion. The mapping of numbers to emotions is typically as follows: 1 for happiness, 2 for sadness, 3 for neutral (no strong emotion displayed), 4 for anger, 5 for surprise, 6 for disgust, and 7 for fear. This numerical labelling system streamlines the process of training supervised learning models by providing a clear, structured format for the emotion data. 

Furthermore, there’s a new way to determine the emotion which is valence and arousal. This way gives each emotion two numbers between -1 to 1. These two numbers represent two axes. Valence represents the emotion is positive or negative, whereas arousal refers to its intensity, from low to high. For example, an emotion with high arousal and positive valence will be happy, an emotion with low arousal and negative valence will be sad. Based on the value of valence and arousal, we can classify the 7 basic emotion states out. 

\subsection{ResNet}
Deep convolutional neural networks have led to a series of breakthroughs for image classification. Deep networks naturally integrate low / mid / high level features and classifiers in an end-to-end multilayer fashion, and the “levels” of features can be enriched by the number of stacked layers (depth). Driven by the significance of depth, a question arises: Is learning better networks as easy as stacking more layers? An obstacle to answering this question was the notorious problem of vanishing/exploding gradients, which hamper convergence from the beginning \cite{he2016deep}.

ResNet provide a brilliant method to prevent vanishing/exploding gradients with deep convolution neural networks. 

To integrate the learned residual g(x) with the input x, ResNet uses a shortcut connection (also known as a skip connection) that bypasses one or more layers. The output of the residual block is then f(x) = g(x) + x, where g(x) is the output from the learned residual function and x is the original input. This mechanism allows the network to adjust the magnitude of the residual, and if the input is already close to the desired output, the network can effectively push g(x) towards 0, making f(x)$\approx$ x. 

\begin{figure}
    \centering
    \includegraphics[width=0.5\linewidth]{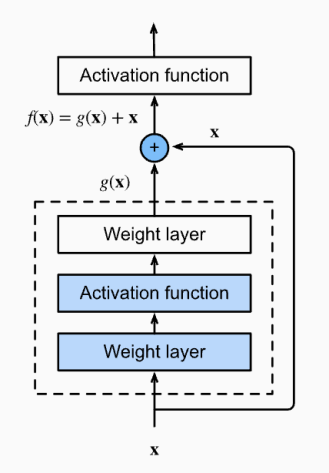}
    \caption{\textit{Figure 2 Architecture of Residual Block }}
\end{figure}

This residual learning framework has a profound implication: it allows the neural network to be significantly deeper without the training becoming unmanageable due to vanishing or exploding gradients. If g(x) is approximately to 0, the f(x) is approximated, indicating that the added layers do not contribute significant changes to the input. This capability not only facilitates the training of much deeper networks by effectively alleviating the vanishing gradient problem but also ensuring that the additional layers do not degrade the network's performance.
Since it publishes, ResNet has become a foundational architecture for many computer vision tasks, setting new benchmarks for performance and efficiency. Its principle of learning residual functions has been adapted and extended to various other types of neural networks, illustrating the versatility and impact of this approach.

\subsection{DenseNet}
DenseNet introduces direct connections between any two layers with the same feature- map size. The paper showed that DenseNets scale naturally to hundreds of layers, while ex-habiting no optimization difficulties \cite{huang2017densely}. 

Dense Connection is a connection method in the DenseNet network architecture that allows each layer to receive feature maps as input from all previous layers and pass its own learned feature maps to all subsequent layers. 
\begin{figure}
    \centering
    \includegraphics[width=0.5\linewidth]{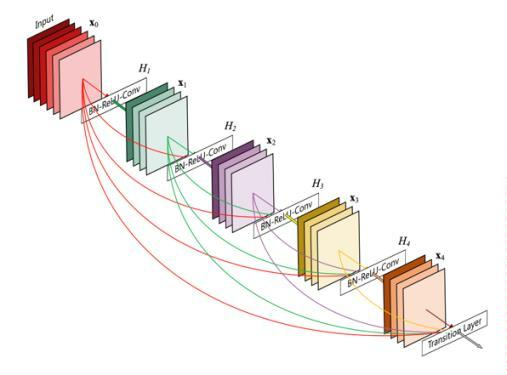}
    \caption{\textit{Figure 3 Architecture of DenseNet }}
\end{figure}
Refer to figure above, if a layer is an X2 layer, it will receive the feature maps of all previous layers X1, X0 as input. This layer performs convolution operations and other operations on the basis of its input feature map to obtain a new feature map. The output feature map of this layer is directly passed to all subsequent layers X3, X4 as part of the input feature map. 

This dense connection method is different from the connection between layers of traditional convolutional networks, where the previous layer is connected to the next layer. Dense connections maximize the reuse of features, and the model can more efficiently combine all features learned in previous layers. So DenseNet enhanced feature propagation helps to better propagate gradients, thereby alleviating the vanishing gradient problem. For the same computing resources, narrower and deeper networks can be built to achieve better accuracy with it. 

\subsection{What Have Been Done}
Classifying correct emotions from different data sources such as text, images, videos, and speech has been an area of research for researchers in various disciplines. Automatic emotion detection from videos and images is one of the most challenging tasks to analyse using supervised and unsupervised machine learning methods.

In a paper, facial and body features extracted by the OpenPose tool have been used for detecting basic 6, 7 and 9 emotions from videos and images by a novel deep neural network framework which combines the Gaussian mixture model with CNN, LSTM and Transformer to generate the CNN-LSTM model and CNN-Transformer model with and without Gaussian centres \cite{karatay2024cnn}. This paper using two benchmark datasets, namely FABO and CK+. It reported over 90\% accuracy for most combinations of features for both datasets.

Another paper, In this paper, therefore, we report the multi-view emotional expressions dataset (MEED) using 2D pose estimation \cite{zhang2023multi}. They got twenty-two actors presented six emotional (anger, disgust, fear, happiness, sadness, surprise) and neutral body movements from three viewpoints (left, front, right). A total of 4102 videos were captured.

A paper introduce emotional body gestures as a component of what is commonly
known as ”body language” and comment general aspects as gender differences and culture dependence \cite{noroozi2018survey}. They also discuss multi-modal approaches that combine speech or face with body gestures for improved emotion recognition.

Together, these studies illustrate the dynamic evolution of emotion detection technologies, showcasing a trend towards more integrated, multi-modal systems that can adapt to and accurately interpret the complex interplay of human expressions and gestures across diverse settings.

\subsection{\textbf{Summary }}
In order to comprehend human emotions, emotion recognition has developed into a sophisticated field that integrates multiple technologies. Due to the direct display of emotions in speech and facial expressions, traditional emotion identification techniques mostly use these methods. Furthermore, because physiological signals reliably indicate subconscious emotional states, such as EEG, ECG, GSR, and ET, they are used to quantify emotional responses. 

Despite advancements in recognizing emotions from facial and physiological signals cues, the role of body gestures remains less explored. Recent studies, like the use of 2D pose estimation, have demonstrated that body gestures can sometimes express emotional information more effectively than facial expressions or speech. OpenPose has notably advanced this area by enabling the analysis of body gestures, which is critical for emotion recognition, especially in settings where facial expressions are not visible. 

The application of machine learning and deep learning has been critical in enhancing emotion recognition systems. Techniques such as Support Vector Machines (SVM) and Convolutional Neural Networks (CNN) are particularly effective in processing and classifying image data. The introduction of ResNet has further improved deep learning models by mitigating the gradient vanishing problem, thus facilitating the training of deeper neural networks. 

Important contributions have also been made by the specialised databases DFEW and Aff-Wild2, which provide extensive video datasets captured in videos and movies. These resources are invaluable for training and validating emotion recognition algorithms under realistic conditions. The classification of emotions into 7 discrete categories such as anger, disgust, and happiness, in other hand newer methods that assess emotional valence and arousal, provides an extensive framework for understanding and interpreting human emotions. 

In general, the field of emotion recognition is moving towards more integrated, multimodal systems that use traditional and innovative methods to capture the subtle and complex aspects of human emotions. This will enhancing applications across various domains including interactive gaming, psychological analysis, assistive technologies, and security systems.

\section{\textbf{Methodology }}
\subsection{\textbf{OpenPose }}
OpenPose is a versatile computer vision tool that specializes in real time multiple person keypoint detection. This technology can analyse videos to identify human body, face, hand, and foot keypoints, providing detailed positional data in keypoints for each frame. The results from OpenPose can be outputted in two distinct formats: images and JSON files. 

In the case of the image format, OpenPose generates image where the detected keypoints are highlighted with coloured points and lines on a black background. This approach focuses only on the keypoints, discarding the original background and other non-keypoint elements of the video frame. This method helps in enhancing the keypoints data, making it easier to analyse face expression and body gesture without the distraction of the original frame's content. 

For the JSON format output, OpenPose provides a structured data file containing detailed information about the keypoints, including their coordinates in each video frame. This JSON output is particularly useful for further computational analysis and can be seamlessly integrated as input into various neural network models. 

\subsection{\textbf{Pre-processing }}
Initially, video files are processed using OpenPose, it will extract human face and body keypoints from video frames. This results in each video being broken down into a series of individual frame images, stored in separate folders corresponding to each video name. 

Two different datasets are involved, each providing numerical labels that correspond to seven different emotions expressed in the videos. Based on these labels, the frame images from each video are then categorized into one of seven designated folders. Each folder represents one of the seven emotions being trained. 

Once the images are sorted into these emotion-specific folders, it's important to locate the issue of redundancy of the data. Because the original video frames are captured in quick succession, consecutive frames often display very similar gestures and expressions due to the continuous nature of video recording. This high similarity between nearby frames can lead to redundancy, which is not ideal for training machine learning models that perform better with diverse data samples. 

To mitigate this, a subsampling technique need to be utilized. Specifically, I chose to retain only every tenth frame from each video. This approach significantly reduces the total number of images while ensuring that the remaining frames are spaced out enough to capture varied gestures and face expressions, thus maintaining the diversity necessary for effective model training. 

By restructuring the data in this way—organizing it into emotion-specific categories and reducing redundancy through subsampling—the pre-processed data is better suited for subsequent analytical tasks. 

After the subsampling, frame images will be categorized to seven different emotion folders based on the labels. 

\subsection{\textbf{Inputting Datasets }}
For importing the images, start by defining the location of the training dataset and the validation dataset. Utilize the dataset.ImageFolder() function to import these datasets while applying data augmentation techniques simultaneously such as rotation, horizontal flip, and normalization. This approach not only simplifies the process of handling dataset but also enhances the dataset by introducing variability and improving model generalization capabilities. 

For importing the JSON files, particularly those containing keypoints data, the initial step is to examine the JSON structure to locate the 'people' key. Subsequently, focus on extracting 'pose\_keypoints\_2d' and 'face\_keypoints\_2d' from the 'people' key. These keys contain valuable data on the positional coordinates of body and facial keypoints, which are essential for tasks that require precise human gestures and facial expression analysis. 

\subsection{\textbf{Showing Images }}
To effectively manage and inspect a dataset of images labeled with emotional expressions, it is beneficial to display the images alongside their labels using two distinct labeling systems. 

The first system categorizes 7 emotions into one of seven basic categories from label 0-6. This categorical system simplifies the complexity of human emotions into easily 7 distinguishable groups, useful for basic emotional analysis. 

The second labeling system employs a two-dimensional approach where each image is associated with two values which are valence and arousal. Valence measures the pleasantness of the emotion, ranging from negative to positive, whereas arousal measures the intensity of the emotion, from low to high. This system provides a more nuanced understanding of emotional states, capturing the subtleties in how emotions are experienced. 

When importing these datasets, it is critical to visually inspect the images alongside their labels to check if the labeling correct and ensure that there are no discrepancies or errors in the dataset and label. Any mismatches identified during this inspection can then be addressed to improve the reliability and usability of the dataset for further emotional analysis. 

\subsection{\textbf{Defining Model }}
In this project, I used three types of neural network architectures: Artificial Neural Network (ANN), ResNet18 with a modified structure and ResNet18 pretrained on ImageNet, and a pre-trained ResNet50 model trained on ImageNet. 

For the ANN, I designed the architecture to include four linear fully connected layers. These layers were connected using ReLU (Rectified Linear Unit) activation functions, which import non-linear function into the network and help prevent the vanishing gradient problem during training. The purpose of this design was to ensure a robust feature extraction from the input data, culminating in the final layer. This last layer is crucial as it outputs the probabilities across seven different emotions, hence the output channel was set to seven. This setup is particularly aimed at facilitating the model's capability to perform multi-class classification of emotional states from the input data. 

Moving to the ResNet18 architecture, I implemented a variant of the standard model to better suit the specific requirements. Each residual block of the network contains two convolutional layers. These convolutional layers are followed by batch normalization layers, which stabilize and accelerate the training process by normalizing the inputs to each layer. A key feature of ResNet architectures is the inclusion of skip connections that help in alleviating the problem of training deep networks by allowing gradients to flow through the network without significant vanishing. The network structure is completed by sequentially connecting these residual blocks with additional convolutional layers and a classification layer, which collectively function to effectively capture and interpret the hierarchical features of the data. 

Lastly, for the ResNet50 model, I utilized a pre-trained version available through the PyTorch 'models' module by executing the following command: \\ models.resnet50(pretrained=True). This model is initially trained on ImageNet, a large visual database designed for use in visual object recognition software research. By utilizing a pre-trained model, we can use the learned weights as a starting point, thus saving significant training time and computational resources. To fits this model to my task of emotion prediction, I modified the final layer to have an output size of seven to correspond to my seven emotion categories. This adaptation allows the ResNet50 model to output refined predictions specifically for my dataset. 

\subsubsection{\textbf{Hyperparameter }}
In the process of fine-tuning the hyperparameters for different machine learning tasks, distinct strategies were employed for classification and regression problems, primarily concerning the choice of loss functions. For the classification tasks, I opted to use the CrossEntropy loss function, which is well-suited for problems involving multiple classes as it effectively handles the probabilistic interpretation of class predictions. Here’s a formula of CrossEntropy below: 

\begin{equation}
    H(p, q) = - \sum_{x=\text{classes}} (p(x) \log q(x))
\end{equation}

H represents CrossEntropy, p is real label of classes while q is the predict output of this classes. Each predicted class will compare to real label of 0 or 1. The calculated loss penalizes the probability based on how far it is from the expected value. Then the penalty is logarithmic, which means there’s a large score for more difference close to 1 and small score for less difference close to 0. 

Conversely, for regression tasks, the Mean Squared Error (MSE) loss was utilized, which quantifies the average of the squares of the errors—that is, the average squared difference between the estimated values and the actual value. Here’s a formula of MSELoss below: 

\begin{equation}
    MSE = \frac{1}{N} \sum_{i=1}^{N} (y_i - \hat{y}_i)^2
\end{equation}

It calculates the squared difference between the predicted value and the actual value. Then then averages these squared differences across N which is number of samples. MSE is great for ensuring that the model we train does not have outlier predictions with huge errors because MSE gives more weight to these errors due to the squared part of the function. 

Adjustments to the learning rate followed the selection of the loss function, as this parameter critically influences the convergence speed and quality of the training process. An inappropriate learning rate can either lead to very slow convergence or cause the training to oscillate without stabilizing. 

Additionally, the choice of optimizer played a crucial role in the network's performance. Various optimizers were tested, including Stochastic Gradient Descent (SGD), Adam, and Adamax. Among these, Adam yielded the best results, likely due to its adaptive learning rate capabilities, which help in handling sparse gradients on noisy problems. 

Adam is an algorithm for first-order gradient-based optimization of stochastic objective functions, based on adaptive estimates of lower-order moments. The method is straightforward to implement, is computationally efficient, has little memory requirements, is invariant to diagonal rescaling of the gradients, and is well suited for problems that are large in terms of data and/or parameters. Furthermore, AdaMax is a variant of Adam based on the infinity norm \cite{kingma2014adam}.

Lastly, learning rate schedulers were implemented to further optimize the training process. Specifically, the OneCycleLR and ReduceLROnPlateau strategies were tried. OneCycleLR aims to adjust the learning rate according to a one-cycle policy that oscillates between a lower bound and an upper bound, potentially leading to faster convergence. The ReduceLROnPlateau scheduler decreases the learning rate when a metric has stopped improving, which ideally reduces overfitting and leads to a more robust model. However, these schedulers did not result in significant improvements in the context of the tasks, suggesting that the benefits of these approaches might be context-dependent or require further adjustment to the hyperparameters. 

\section{\textbf{Implementation }}
\textbf{Artificial Neural Network (ANN) }

The artificial neural network (ANN) I implemented consists of a simple architecture with four linear layers. The network is designed to process input data provided in a JSON file, which mainly contains keypoints. 

As data goes through the network, it through four different linear layers. The main goal of these layers is to classify input data into one of seven basic emotional states. This classification is based on patterns and relationships learned by the network from training data involving emotional cues in the keypoints. 

In addition to classifying emotional states, the network can also evaluates valence and arousal, two dimensions commonly used to describe emotions. To achieve this, the input data is first passed through three separate linear layers dedicated to processing these specific features. Thereafter, the processed data are further refined through two additional linear layers. The output of these two layers is simply output a value for each layer representing valence and arousal. The specific task of these layers is to predict valence and arousal values based on input keypoints. 

Using multiple linear layers in this way enables artificial neural networks to learn from complex high-dimensional data sets and make accurate predictions of emotional state, valence, and arousal. This structured approach enables the network to effectively differentiate between different emotions and assess their intensity and pleasantness, providing valuable insights for applications requiring emotion recognition. 

\textbf{Residual Neural Network (ResNet) }
\begin{figure}
    \centering
    \includegraphics[width=1.\linewidth]{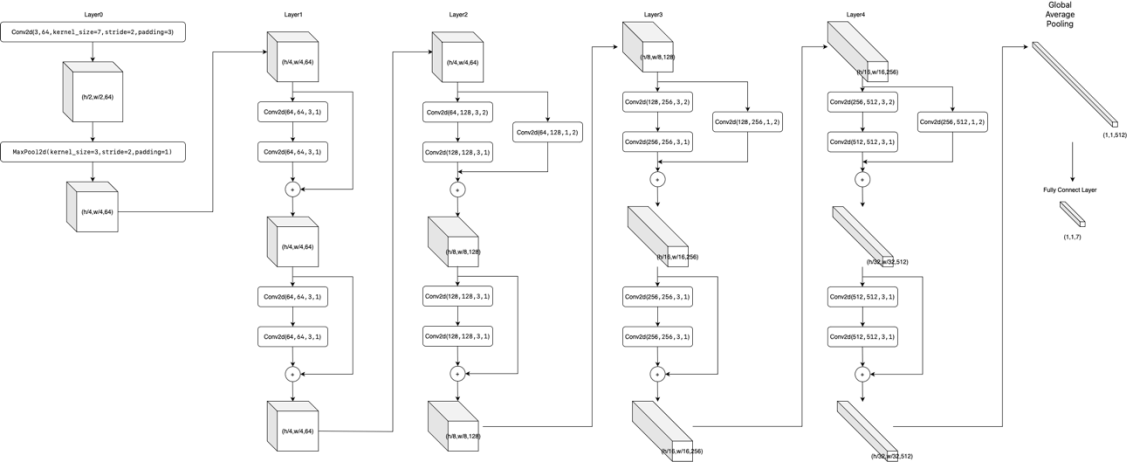}
    \caption{\textit{Figure 4 Detailed Architecture of ResNet18 }}
\end{figure}

The architecture I've implemented known as ResNet18. Comprises a total of 18 distinct layers designed to enhance the model's performance and efficiency in processing visual data. The initial layer, the Layer0, is a convolutional layer that serves as the entry point for the input images. This layer is crucial as it performs the initial processing and transformation of the input. 

Between the initial and final layers, the model is structured into four main layers—Layer1 through Layer4. Each of these layers contains two residual blocks, making up the core mechanism that enables ResNet models to train more effectively and avoid issues related to deep networks such as vanishing gradients. A residual block is specially designed to allow activations to propagate through the network more effectively by incorporating skip connections that skip two layers in my model. 

Within each residual block from Layer1 to Layer4, there are two convolutional layers. These layers are responsible for extracting and refining features from the input data, with each subsequent layer building on the output of the previous layer. The design of these blocks and their shortcut connections help reduce information loss during training, ensuring that the network learns effectively even as it deepens. 

Following Layer4, the architecture includes a fully connected layer, which typically comes towards the end of the network. This layer is essential for integrating and interpreting the features extracted by the convolutional layers, making high-level decisions about the input data based on the learned features. In classification task, it should be 7 which represent 7 basic emotion states. Then for valence and arousal, the output channel of fully connected layer should be 2 which are values of valence and arousal. 

Summing up all the convolutional layers from Layer1 to Layer4 results in 16 convolutional layers. When combined with the initial Layer0 and the fully connected layer, the total comes to 18 layers, hence the name ResNet18. This architecture is known for its robustness and efficacy in handling complex visual tasks by leveraging deep learning capabilities while maintaining manageable computational costs. 

For ResNet50, which is quite similar with ResNet18. The difference is ResNet50 get deeper and more layers. ResNet50 use more residual blocks which means more convolutional layers. But it also requests higher computational resource and has a bigger model size. Due to this, it generally achieves better performance on both classify and regression tasks due to its increased depth and capacity to learn more intricate features. So, it will get better result than ResNet18. 

For DenseNet121, compared to ResNet18 and ResNet50, DenseNet121 is a much deeper network, with more layers and a different connectivity pattern. While ResNet use skip connections to mitigate the vanishing gradient problem, DenseNet employ dense connectivity, which has proven to be an effective way to train very deep neural networks. In terms of performance, DenseNet121 generally outperforms ResNet18 and ResNet50 in my project. 

\section{\textbf{Result }}
ANN use JOSN as input, JSON got more accuracy and as a text file it need less computational resources than images. As the result, ANN needs less time for training. For the other neural network, they use images as input. Images are suitable for CNN which designed for image data. And visual representation of the image helps me quickly verify the results. 

\subsection{Highest F1 Score for Classification with 7 Emotion States }

\begin{table}[h]
    \centering
    \begin{tabular}{ccc}
        Model &  SGD& Adam\\
         ANN(JSON)&  0.18& 0.17\\
         ResNet18&  0.26& 0.29\\
         ResNet50&  0.14& 0.32\\
         DenceNet121&  N/A& 0.35\\
    \end{tabular}
\end{table}
The F1 score is a measure of a model's performance, commonly used in fields like machine learning and information retrieval. It combines the precision and recall metrics into a single score. Precision measures how many of the predicted positive cases were correct. Recall measures how many of the actual positive cases were correctly identified. Then use F1 = 2 * (precision * recall) / (precision + recall) to calculate the F1 score. 

\subsection{Highest CCC Score for Valence and Arousal Task }

\begin{table}[h]
    \centering
    \begin{tabular}{ccc}
         &  SGD& Adam\\
         ANN(JSON)&  1.72e-11& 5.04e-12\\
         ResNet18&  0.18& 0.26\\
    \end{tabular}

\end{table}
The CCC (Concordance Correlation Coefficient) score is a metric used to evaluate the performance of regression models by measuring the agreement between predicted and real values. 

\begin{equation}
    CCC = \frac{2 \times \text{covariance}(x, y)}{\sigma_x^2 + \sigma_y^2 + (\mu_x + \mu_y)^2}
\end{equation}

$\sigma ^2$ means variance and $\mu$ means mean in this equation. 

In general, deeper model get better result. And Adam always gets best result faster than SGD. And Adam usually gets better result because Adam can adjust the learning rates based on the average of the second moments of the gradients. So the pre-trained DenseNet121 get the best result. Lastly the effectiveness of SGD can be notably enhanced by fine-tuning the learning rate. So sometimes SGD can get better result by adjusting learning rate. 

\section{\textbf{Evaluation }}
\subsection{\textbf{Strength of Projects }}
The project use OpenPose, an advanced tool capable of extracting keypoints from videos of individual person, OpenPose will focus exclusively on the body and facial keypoints while discarding other potentially distracting elements like the original image. The extraction is done in two distinct ways: firstly, by rendering coloured keypoints over a plain black background in an image format, and secondly, by saving the keypoints' coordinates in JSON file. These two approaches not only streamline the data extracting but also enhances processing efficiency. 

To enhance the richness of the dataset, I implement two large datasets, Aff-Wild2 and DFEW, which provide a total of approximately 4,000,000 video frames covering sufficient variation in the seven basic emotional states. Such a large amount of data ensures a comprehensive sampling of human emotional expressions, thereby facilitating robust model training. 

In the pre-processing phase, I employed techniques such as subsampling and classification of the images into seven distinct categories. This methodological categorization helps minimize redundancy and prevents potential I/O errors when handling large volumes of image and JSON data. Following preprocessing, I utilized the images with coloured keypoints for training a convolutional neural network using ResNet architectures (specifically ResNet18 and ResNet50). For the JSON data, a three-layer artificial neural network was used. By comparing the effectiveness of two optimizers, SGD and Adam, within the context of the Cross Entropy(classify) and MSE (regression) loss function, it was determined that the Adam optimizer exhibited superior capability in feature learning, hence optimizing our neural network training. 

\subsection{\textbf{Weakness of Projects }}
The project is reliance on OpenPose brings inherent limitations, particularly in terms of accuracy and completeness of extracted keypoints. Sometimes OpenPose is unable to detect certain parts of the body or face, resulting in an incomplete representation of the data, such as frames with complete loss of key keypoints resulting in a completely black picture. 

Processing video frame by frame using OpenPose also results in a quite high similarity between consecutive frames. This similarity poses a significant challenge in maintaining the diversity of the training dataset, as many frames do not differ enough from one another. To alleviate this problem, I implemented secondary sampling, selecting every ten frames to reduce redundancy and enhance dataset diversity. However, challenges remain even after secondary sampling, mainly because many of the frames derived from successive video sequences contain transitional or less expressive emotional states that are less obvious and therefore less useful for training. 

The project initially divided emotions into seven categories, consistent with traditional seven emotion states. However, due to the consecutive frames cause highly similarity between nearby frames and due to OpenPose there’re lots of invalid frames. To address these challenges, I turned my focus to using valence and arousal as a dimensional approach to continuous output, turning the problem into a regression task. This change allows the network to predict two values, which are then used to set thresholds for seven basic emotions, providing a potentially more nuanced understanding of emotional states than just categorical categories.

\subsection{\textbf{Summary of Evaluation }}
According to the results from model, the main issue is low accuracy. That’s because high similarity among consecutive frames, processing video frames sequentially with OpenPose resulted in a high degree of similarity among them, which hindered the diversity of the training dataset. Despite attempts to mitigate this issue through subsampling (selecting every tenth frame), the problem persisted. Many frames did not sufficiently differ from each other, particularly those depicting transitional or less expressive emotional states. 

Initially, the project categorized emotions into seven discrete classes based on traditional emotion theory. This approach faces difficulties due to the indistinguishable and large proportion of emotion transitions between frames, resulting in low verification accuracy. So many similar images in different classes make the ResNet and DenseNet model cannot identify the difference, making it difficult to differentiate and correctly classify the emotions. 

To solve this problem, I change my model from a classification task to a regression task, categorizing emotions based on valence and arousal as continuous outputs. This shift aimed to enhance the model's capability to capture a more accurate and more nuanced spectrum of emotional expressions. However, the there’re lots of similar images, it still potentially improve the accuracy. 

\section{\textbf{Conclusion }}
This project provided a good opportunity to use my background in computer science to dive into new areas across disciplines. Through this initiative, I was able to take a hands-on approach to the development of computer vision technology, methodically addressing challenges at each step of the process. As the project progressed, my passion for the area of computer vision grew further. 

The main goal of the project is to develop a predictive model capable of recognizing seven different emotions: happiness, sadness, anger, fear, surprise, disgust and neutral states. I use OpenPose for image processing to extract features required for emotion recognition. The method first pre-processes the images and then organizes them into seven different folders corresponding to each emotion to facilitate the classification task. I also explore subsampling techniques to enhance the model's ability to handle regression tasks. However, due to the nature of the video-derived dataset (which mainly consists of sequential images), the sample lacks diversity even after a subsampling effort. 

In the comparative analysis, I found that ResNet and DenseNet get better performance than traditional artificial neural networks (ANN) on both regression and classification tasks. Performance differences can be attributed to different datasets and structural differences in the models. In the classification task, it is clear that the model can more accurately predict certain emotions, such as "neutral" and "happiness." This may be because these emotions exhibit unique and consistent cues that are easier for models to find out. In contrast, detect emotions such as fear and anger pose more significant difficult due to subtle differences in expression. 

Furthermore, when performance was assessed in terms of valence (the intrinsic attractiveness or aversiveness of an event) and arousal (the physiological and psychological state of being aroused or aroused), it was observed that these models generally exhibited lower loss rates, suggesting a greater sensitivity to emotional intensity. 

Overall, the application of ResNet and DenseNet in this project proved they are very effective for emotion recognition tasks. The use of the Adam optimizer enhances the performance of them, achieving the best results in both regression and classification of emotional states. I think that by adjusting the diversity of training data and further tuning model parameters, the accuracy of valence and arousal predictions may improve. 

\subsection{\textbf{Further Work }}
Possible further work could implement advanced subsampling techniques to selectively retain representative images from sequences of continuous actions., in order to reduce the similarity of images between different classes. 
Possible is using a better computer which will provide a better GPU or multiple GPUs rather than single GPU, can significantly reduce training and inference times. This could also include exploring cloud-based solutions or specialized hardware like TPU. 
Possible to use another CNN, like DenseNet201, which features densely connected convolutional networks, or Vision Transformer (ViT), which applies transformer mechanisms to image recognition, might yield better or faster results. Implementing techniques like grid search, random search, or Bayesian optimization to automate the selection of model hyperparameters, which could enhance model performance without extensive massive manual experimentation. 

The proposed directions for further research in project systems focus on three areas to overcome the weakness of project. 

First, optimizing data handling, for example, subsampling gets developing more efficient methods for processing and managing data, which can lead to faster response time and reduce cost of computational resource. This might include the implementation of advanced data subsampling or the use of more complex data pre-processing techniques to enhance the quality of the data. 

Second, experimenting with more advance neural network architectures is crucial for improving the accuracy of this system. This might involve utilizing new forms of deep learning models, such as recurrent neural networks (RNN) or Generative Pre-trained Transformer (GPT), By modifying existing architectures or creating new ones, model can be better capturing the nuances of emotional expressions, potentially leading to breakthroughs with this project. 

Finally, these strategies aim to optimize the performance of project by making them faster, more accurate, and capable of processing larger and more diverse datasets. These improvements are particularly important for applications that demand real-time processing and high levels of precision, such as customer service analysis, and healthcare monitoring, where rapid and reliable emotion detection is crucial.

\section{Acknowledgements}
A special thank you to my supervisor Dr Dimitrios Kollias, for his guidance, expertise, and encouragement throughout the project.

I wish to extend a special thanks to my parents and my little sister for the constant love, inspiration, and support throughout. Which enabled me to dedicate my time and effort to this project. As well as my friends for continuous solace and motivation.

\bibliographystyle{ieeetr}
\bibliography{references}

\end{document}